\documentclass[final,5p,times,twocolumn]{elsarticle}

\usepackage{amssymb}
\usepackage{amsmath}

\usepackage{enumitem}
\usepackage{mathtools}
\usepackage{lipsum}
\usepackage{multirow}
\usepackage{makecell}
\usepackage{array}
\usepackage{booktabs}
\usepackage{mathrsfs}
\usepackage{xcolor}
\usepackage[table]{xcolor}
\usepackage{colortbl}
\usepackage[colorlinks=true, linkcolor=blue, urlcolor=blue, citecolor=blue]{hyperref}
\usepackage{ulem}

\journal{Pattern Recognition Letters}

\begin{document}

\begin{frontmatter}

\title{FSE: Continual Learning for Named Entity Recognition by Fast-Slow Experts}

\author[label1]{Yunan Zhang}
\ead{zhangyunan@stu.hit.edu.cn}
\author[label1]{Yang Fan}
\ead{yfan@stu.hit.edu.cn}
\author[label1]{Heng Li}
\ead{20b951012@stu.hit.edu.cn}
\author[label1]{Xiangping Wu\corref{cor1}}
\ead{wuxiangping@hit.edu.cn}
\author[label1]{Qingcai Chen\corref{cor1}}
\ead{qingcai.chen@hit.edu.cn}

\affiliation[label1]{organization={School of Computer Science and Technology, Harbin Institute of Technology},
                    city={Shenzhen},
                    state={Guangdong},
                    country={China}
}

\cortext[cor1]{Co-corresponding authors}

\begin{abstract}
Continual Learning for Named Entity Recognition (CLNER) enable models to incrementally learn new entity types without forgetting previously acquired ones. However, existing methods suffer from catastrophic forgetting and insufficient exploitation of shared information across tasks. This paper proposes \textbf{FSE}, a \textbf{F}ast-\textbf{S}low \textbf{E}xperts enhanced span-based NER model for CLNER. The shared fast expert learns token-level links to efficiently filter out unlikely spans, while the task-specific slow expert performs span classification only on the remaining candidates. It stabilizes learning by promoting knowledge sharing across tasks and maintains plasticity by reducing learning burden at each task. A length-decay negative sampling strategy to mitigate span imbalance is also introduced. Extensive experiments on OntoNotes and FewNERD synthestic datasets demonstrate that FSE achieves state-of-the-art performance in CLNER scenarios, with effectiveness of each component, empirical evidence of faster convergence and expected functionality of both experts.
\end{abstract}



\end{frontmatter}

\section{Introduction}\label{sec:intro}
Continual learning is pivotal for advancing evolvable artificial general intelligence, enabling models to acquire new knowledge from streaming data~\cite{wang2024comprehensive}.
In this field, continual learning for named entity recognition (CLNER) is key to developing real-time or personal application that requires dynamic adaptation to changing scenarios.
NER aims to extract entities from unstructured text, CLNER further enables the incremental learning of new entity types over time without forgetting previously learned ones~\cite{monaikul2021continual}.
For example, after learning to detect medical terminology by training on medical text, the model can still accurately identify previously learned entity types like persons and locations, without requiring retraining from scratch.

CLNER always encounters catastrophic forgetting~\cite{Benk2024ExampleFA,Zhou2024PNSPOC}, where models trained on new tasks tends to rapidly disrupt previously learned ones. To mitigate this, standard NER models often incorporate anti-forgetting techniques such as knowledge distillation or model expansion. Meanwhile, these efforts also witness the paradigm shift of NER from sequence tagging to span-based methods~\cite{Seow2025ARO}.
Here we focus on span-based methods, which enumerates and classify all possible text spans, providing richer representations and often yielding superior performance.
Notably, recent large language models (LLMs) show promise in CLNER owing to their strong generalization capabilities. However, empirical studies~\cite{liu2025concept} reveal that LLMs still underperform their smaller, NER-specific Bert-based counterparts. Moreover, The prohibitive computational demands and challenges in dynamic updates of large parameters also hinder its practical deployment.

\begin{figure}
\centering
  \includegraphics[width=0.9\columnwidth]{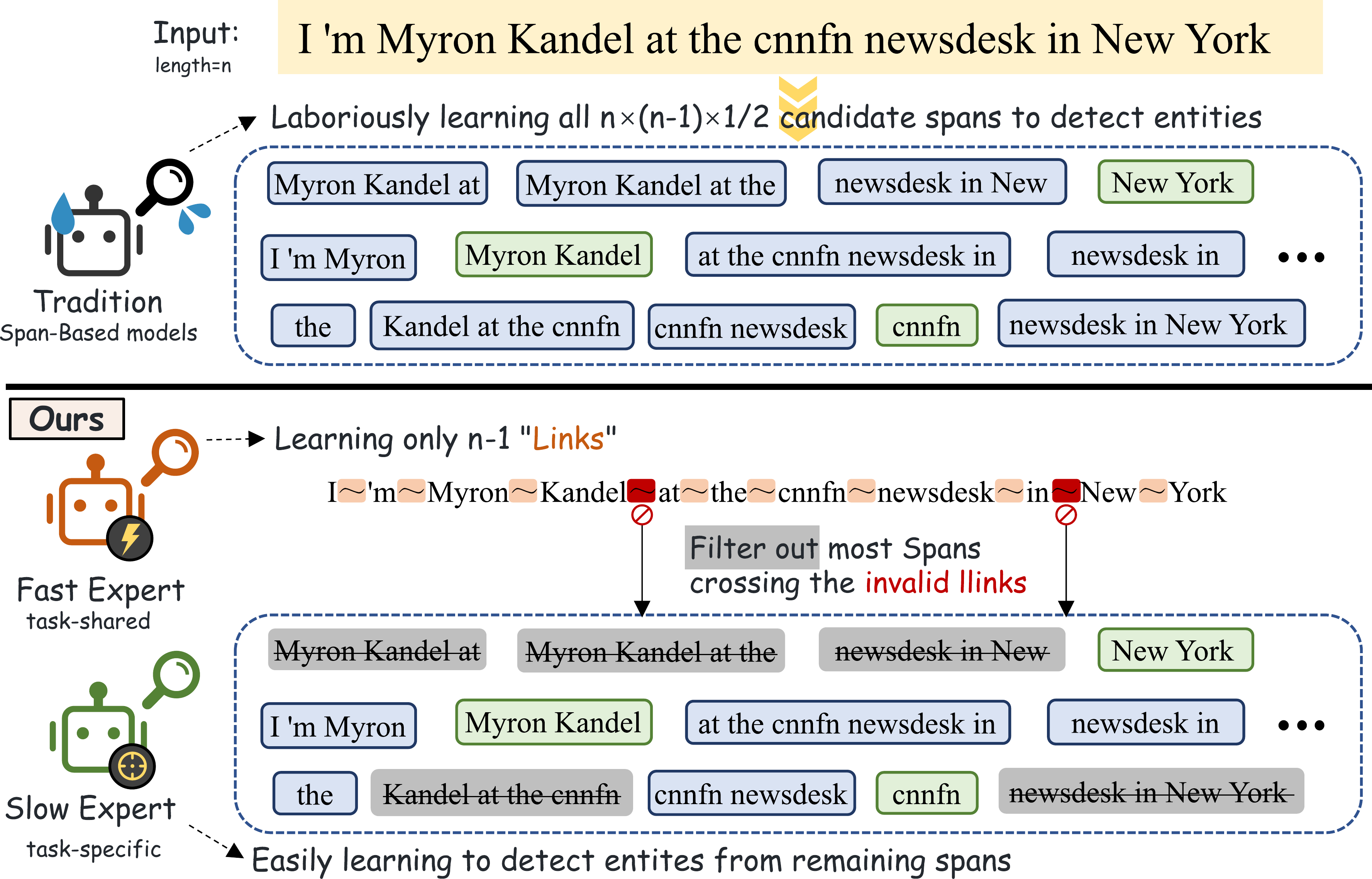} 
  \caption{Compared to learning massive spans in traditional methods, the fast expert in FSE efficiently learns a few links to filter out most spans, which reduces the learning burden on the slow expert that ultimately detects entities.
  }
\label{introduce}
\vspace{-15pt}
\end{figure}

Hence, span-based CLNER methods remain preferable, but we argue that there still exist unsolved challenges from two aspects:
\textbf{1) First,} task interference may lead to conflicting parameter updates during continual learning. Thus, models capable of explicitly modeling task-shared information can facilitate stability in this process. However, existing span-based methods, which only share the Bert base encoder, fail to explore additional high-level information specific to NER that could be shared between different tasks. For example, when a span text "\texttt{[PER] at}" (a person entity but followed by a preposition) has already been learned as a non-entity span in previous tasks, this pattern is also unlikely to be the entity in subsequent tasks. Such knowledge could be shared. \textbf{2) Second}, model's plasticity gradually diminishes as learning progresses~\cite{Dohare2024LossOP}, making the acquisition of new entity types increasingly difficult. Thus, reducing the learning burden for each task would benefit continual learning. For example, given a currently learned non-entity span "\texttt{[PER] at}", model should readily identify the extended span "\texttt{[PER] at the}" as a non-entity as well.

Here we propose that a properly designed "links" between adjacent tokens can address both considerations. Specifically, we introduce a \textbf{F}ast-\textbf{S}low \textbf{E}xperts enhanced span-based model (\textbf{FSE}), where the fast expert (shared across tasks) models these links, while the slow experts (task-specific) performs standard span classification at each task.
As shown in Fig~\ref{introduce}, given a sentence with $n$ tokens, traditional span-based model would laboriously enumerate all $0.5(n^2-n)$ spans to detect entities for each task. Instead, our fast expert first learns the relatively fewer links ($n\!-\!1$) between adjacent tokens. We define that links within an entity span are all strong, while non-entity spans must contain at least one weak link. Notably, this link mechanism is merely a designed inductive bias imposed on the model, the specific link strengths are still automatically learned by the model itself during training. Upon this design, we expect that once the fast expert learn a weak links (e.g, between "\texttt{[PER] at}", massive spans cover this link can be filtered out rapidly. Then the remaining fewer spans are focused by the slow expert.
This substantially alleviated the span learning burden of slow expert, and the sharable mechanism of fast experts also benefit the newcome slow experts in subsequent tasks.

In implementation, dual experts are jointly optimized to avoid error propagation. First, the fast expert applies softmin pooling to aggregate the learned link scores within a certain span into this span's "fast score", ensuring the span-level fast score tends to be low if any weak link is detected. Next, The fast score is fused with the slow expert's output score for each span, ensuring that an entity is recognized only when the fast expert confidently identifies strong links and the slow expert confirms it as a valid entity. Spans without meeting both conditions would be rejected.
During continual learning, we use knowledge distillation to distill the fused outputs of dual experts in previous tasks for transferring prior knowledge.
Moreover, benefiting from fast expert's efficiency in filtering out spans, fewer negative spans are needed for explicit learning. This makes negative sampling, a technique aiming to mitigate the positive-negative span imbalance inherent in span-based models, particularly suitable.
We propose a length-decay negative sampling strategy that adjusts the sampling probability based on span length, applying it to both current task learning and old task distillation.

Extensive experiments on synthesized datasets from OntoNotes and FewNERD show the effectiveness of FSE, achieving SoTA performance among various baseline including LLMs methods.
The contributions of this paper include:
\begin{itemize}[leftmargin=*, align=left]
\item We identify the potential of adjacent token links to promote the stability and plasticity in span-based CNLER model.
\item We introduce jointly optimized fast-slow experts to enhance standard span-based CLNER model, where the fast expert utilizes links to guide the span modeling of slow expert. This reduces task-specific learning difficulty while enabling effective inter-task knowledge sharing.
\item Extensive experiments including LLM-based competitor demonstrate superior performance of our FSE in CLNER scenarios.
\end{itemize}

\section{Related Work}\label{sec:related}
\textbf{Named Entity Recognition} constitutes a cornerstone of information extraction in natural language processing (NLP), aiming to recognize predefined entity types from unstructured text.
NER is traditionally formalized as sequence tagging by adopting CRF after neural representations~\cite{lample2016neural,strakova2019neural}.
In contrast, span-based methods enumerate all possible spans within the text and directly classify them without tagging scheme (BIO) conversion and accommodate nested entities~\cite{yu2020named,zhang2023neural}.
Besides, Machine Reading Comprehension (MRC)-based methods use pointer networks to locate entity positions in the input text~\cite{mrcner2020,GFMRC2023}.
Beyond these extraction-based paradigm, generation-based approaches such as Seq2Seq-based directly generate entity mentions from the given input~\cite{yan2021unified,lu2022uie}.
Recent LLM-based NER methods are also generative, exhibiting remarkable generalization capabilities in handling both seen and unseen entity types. However, smaller models remain valuable due to their efficiency and domain-specific adaptability.

\textbf{Continual Learning NER} tackles the evolving scenarios where new entity types emerge sequentially. This progressive shift in data distribution violates the i.i.d. assumption, leading to catastrophic forgetting of previously learned entity types. CLNER aim to balance the retention of past learning with the acquisition of new information, also known as the stability-plasticity dilemma.
Monaikul~\cite{monaikul2021continual} first framed CLNER as class-incremental continual learning and introduce sequence tagging-based AddNER and ExtendNER.
The training data for each sequential task are annotated only with the the entity types currently being learned, and the model is evaluated on data that requires predicting all accumulated learned entity types. AddNER uses an individual head for each new task and ExtendNER uses a single unified head shared across all tasks, both incorporating knowledge distillation.
Then, L\&R~\cite{xia2022learn} and ExtendNER+DLD~\cite{zhang2023decomposing} both build upon ExtendNER to further improve its performance.
SpanKL~\cite{zhang2023neural} first uses span-based modeling and binary classification in CLNER to alleviate conflicts in non-entity types across different tasks. SKD-NER~\cite{chen2023skd} further uses reinforcement learning upon it.
However, all these methods share information solely through the underlying BERT parameters and neglect the potential to leverage other types of information shared across tasks. In contrast, in CV task such as continual semantic segmentation, architecture decomposition methods have explored decomposing the model into task-specific and task-sharing components, where the task-sharing part supports reconciling old and new knowledge simultaneously, with the task-specific part adapting to new tasks~\cite{yuan2024survey}. For example, Representation Compensation Networks~\cite{zhang2022representation} uses structural re-parameterization to decouple CNN modules into two parallel branches for shared and task-specific information. LAG~\cite{yuan2024learning} disentangles semantic features under channel-wise and spatial-level to simultaneously reconcile knowledge inheritance and new-task learning, where semantic-invariant knowledge is modeled as abstract prototypes shared across tasks.
To the best of our knowledge, we are the first to explore shareable information specific to NER under CL setting, especially span-based methods, to improve overall performance across tasks.

\begin{figure*}[t]
\centering
\includegraphics[width=0.9\textwidth]{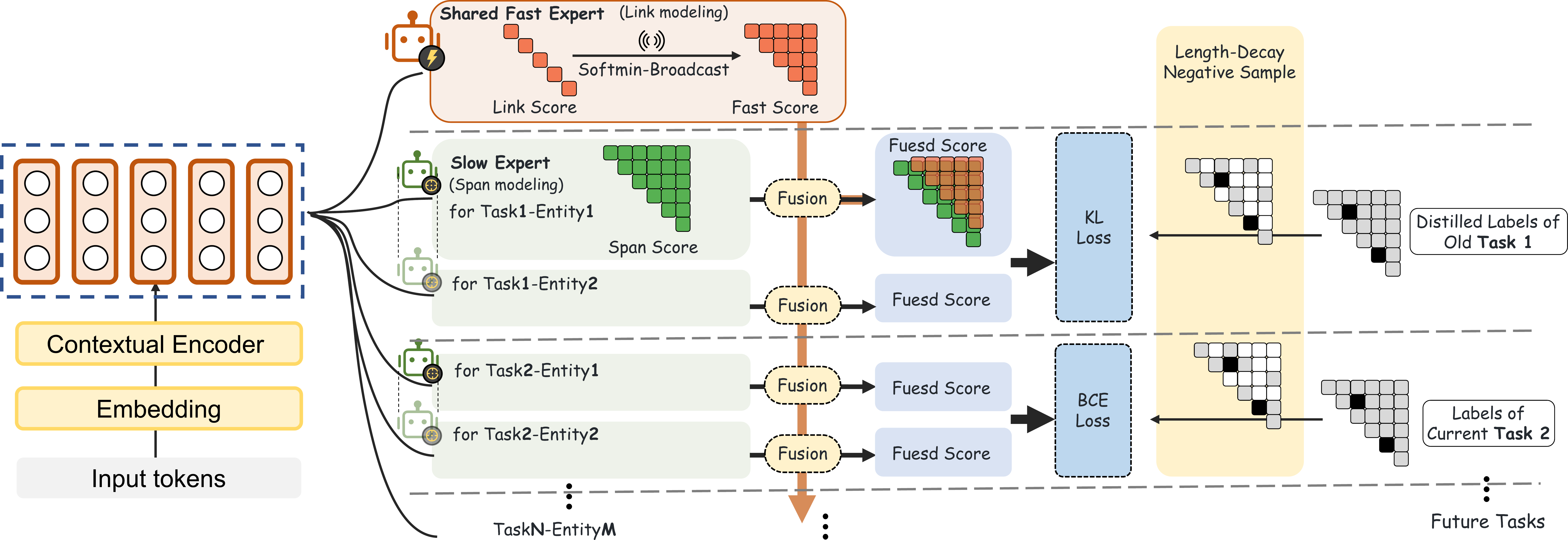}
\caption{Illustration of FSE: The task-shared fast expert for links modeling outputs fast score. The task-specific slow expert for span modeling outputs span score. Both scores are finally fused. Previous tasks and current tasks are optimized by KL loss and BCE loss, respectively. Negative sampling apply in both optimizations.}
\label{model_overall}
\vspace{-10pt}
\end{figure*}

\section{Method}\label{sec:method}
In this section, we first formulate CLNER, and then introduce the base architecture of FSE, including slow expert, fast expert and their collaboration mechanism. Finally, we describe the training techniques employed.

\subsection{CLNER Formulation}
CLNER is formally defined as: An NER model is sequentially trained on a stream of tasks $\mathcal{T}_1, \mathcal{T}_2, \dots, \mathcal{T}_L$. Each task $\mathcal{T}_l$ introduces a new set of entity types $\mathcal{E}_l = \{e_l^1, e_l^2, \dots\}$. Dataset for $\mathcal{T}_l$ only annotates its corresponding entity types $\mathcal{E}_l$. Typically, during training on task $\mathcal{T}_l$, the model cannot access data from previous tasks $\mathcal{T}_1, \dots, \mathcal{T}_{l-1}$.

The learning process begins by training the initial model $\mathcal{M}_1$ on $\mathcal{T}_1$ to recognize entity types in $\mathcal{E}_1$. For each subsequent step $l > 1$, the model $\mathcal{M}_l$ is incrementally trained on the data of $\mathcal{T}_l$ based on $\mathcal{M}_{l-1}$, aiming to recognize all previously learned entity types $\bigcup_{i=1}^{l}\mathcal{E}_i$.

\subsection{Model Architecture}
Fig.~\ref{model_overall} shows the overall architecture of FSE, including contextual encoder for token learning, task-shared fast expert for link learning and task-specific slow expert for span learning.
Negative sampling applies in both KL and BCE optimizations.
\subsubsection{Contextual Encoder}
Given input sentence $X$ with $n$ tokens $[x_1, x_2, ..., x_n]$, we define span $s_
{ij}$ as continuous tokens from $x_i$ to $x_j$, where $1\!\leq\!i\leq\!j\leq\!n$. The contextual encoder capture tokens dependencies and generates contextualized representations for each token. We employ the widely-used Bert-base PLMs as our contextual encoder, formulated as:
\begin{equation}
\mathbf{H} = \mathtt{Encoder}(X)
\end{equation}
where $\mathbf{H} = [\mathbf{h}_1,\mathbf{h}_2,\dots,\mathbf{h}_n]\!\in\!\mathbb{R}^{n \times d^h}$ represents the contextualized representations vector for each token, $d^h$ is hidden size.

\subsubsection{Slow Expert}
The slow expert models the representation of each span to facilitate span classification. We follow SpanKL to use a simple scaled dot-product interaction between the "start" and "end" feature space of token. This means the slow expert is boundary-aware and responsible for detecting span boundaries, enabling the fast expert to focus on the information within the spans. Specifically, for each entity types, we use two distinct single-layer feedforward network (FFN) to yield the start and end representations of tokens, respectively. Then span representation $h^{s_{ij}}$ is computed as:
\begin{equation}
\begin{aligned}
\mathbf{h}^{s_{ij}} & = [h^{s_{ij}}_1, h^{s_{ij}}_2,\dots h^{s_{ij}}_K] \in \mathbb{R}^K\\
h^{s_{ij}}_k & = \mathtt{FFN}^{\text{start}}_k(\mathbf{h}_i)^\intercal \cdot \mathtt{FFN}^{\text{end}}_k(\mathbf{h}_j) \times (d^{o})^{-0.5}
\label{compute_span_startend}
\end{aligned}
\end{equation}
where $k$ denote the k-th entity type among the total K types currently being learned. All the $\mathtt{FFN}^{\text{start}}$ and $\mathtt{FFN}^{\text{end}}$ relative to each entity type have the same output dimension $d^{o}$. There are totally $2\times K$ distinct FFNs. This is similar to multi-head attention mechanism by treating start, end and entity type as Query, Key and Head and enables seamless addition of FFNs initialized for the newcome tasks. 

\subsubsection{Fast Expert}
Let $\alpha_i \in \mathbb{R}$ denote the link score, which quantifies the link strength between adjacent tokens $x_i$ and $x_{i+1}$ in the sequence $X$, structured as:
$x_1,\underline{\alpha_1},x_2,\underline{\alpha_2},x_3,...,x_{n-1},\underline{\alpha_{n-1}},x_n$,
where higher score indicates stronger link. There are total $n\!-\!1$ link scores within sentence of length $n$ and each link score is dynamically computed by the fast expert as:
\begin{gather}
a_i = \mathtt{FFN}^{\text{start}}_{\text{link}}(\mathbf{h}_i)^\intercal \cdot \mathtt{FFN}^{\text{end}}_{\text{link}}(\mathbf{h}_{i+1}) \times (d^{l})^{-0.5} \label{compute_link}
\end{gather}
where $\mathtt{FFN}^{\text{start}}_{\text{link}}$ and $\mathtt{FFN}^{\text{end}}_{\text{link}}$ are two distinct FFNs modeling the link's start and end feature space respectively. $d^{l}$ is output dimension. This process is similar to span modeling in Equ.~\ref{compute_span_startend}, allowing it to be treated as an "additional entity types" modeling step, which facilitates parallel computation in practice.
Notably, $\mathtt{FFN}^{\text{start}}_{\text{link}}$ and $\mathtt{FFN}^{\text{end}}_{\text{link}}$ of fast expert are shared across tasks.

\subsubsection{Span-Level Fast Score}
\begin{figure}
\centering
  \includegraphics[width=0.9\columnwidth]{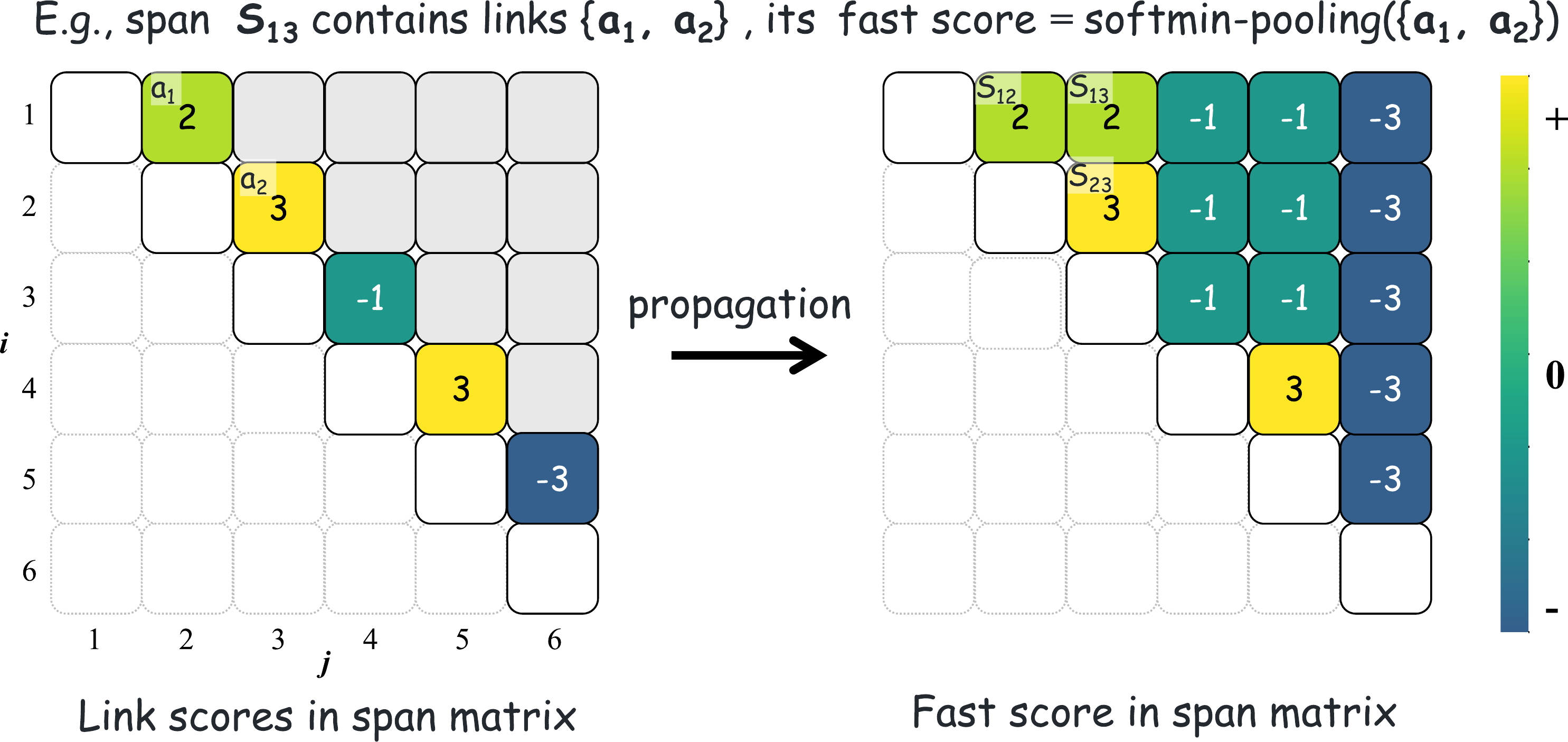}
  \caption{Using softmin-pooling to propagate link scores into span-level fast scores from a matrix perspective.
  }
\label{propagation}
\vspace{-10pt}
\end{figure}

To propagate link information to each span, we need to broadcast the obtained 1D token-level link scores into 2D span-level scores (termed as fast scores $f$).
As shown in Fig.~\ref{propagation}, we aggregate all link scores within a certain span and pooling them into a single relative fast score.
Given our design objective that a span should be rejected if it contains even a single weak link (i,e,. the weakest link acts as an upper bound), we introduce \textit{softmin-pooling}, a stable and smooth version of min-pooling, to mitigate potential gradient sparsity and instability issues. The softmin is implemented via an inverted logsumexp operator as:
\begin{equation}
\begin{aligned}
f^{s_{ij}} &= \text{Pooling}(\{a_i,\dots,a_{j-1}\}) = -\tau \cdot\log\sum_{a \in \mathcal{A}}\exp(-a/\tau)
\end{aligned}
\end{equation}
where $\mathcal{A}\!=\!\{a_i, \dots, a_{j-1}\}$ is the set of link scores within span $s_{ij}$, temperature $\tau$ controls the sharpness of the pooling and is set to $0.1$.
Notably, we also address the challenge of batch-level paralleled implementation for softmin-pooling by heuristically using the \textit{logcumsumexp} operator in PyTorch.

\subsubsection{Dual Experts Fusion}
Given both span-level scores derived from the slow expert: $\mathbf{h}^{s_{ij}}\!\in\!\mathbb{R}^K$ (containing K scalar scores regarding each entity type) and from the fast expert: $f^{s_{ij}}$, we fuse them into final score for classification.
We proactively apply the sigmoid activation to both scores into $0{\sim}1$. As expected, recognition of entity requires consensus from both experts, where a low score from either expert leads to rejection. This fusion requirement can be satisfied by Harmonic Mean. Fused score $\mathbf{\tilde{h}}^{s_{ij}}$ is computed as:
\begin{gather}
h_k^{s_{ij}} \coloneqq \sigma(h_k^{s_{ij}}), \quad f^{s_{ij}} \coloneqq \sigma(f^{s_{ij}}) \label{eq:norm}\\
\mathbf{\tilde{h}}^{s_{ij}}  =\!\textbf{Fuse}(\mathbf{h}^{s_{ij}}, f^{s_{ij}})=\![\text{fuse}(h_1^{s_{ij}}, f^{s_{ij}}), \dots, \text{fuse}(h_K^{s_{ij}}, f^{s_{ij}})] \\
\text{fuse}(x, y) = 2xy /(x + y)
\end{gather}
where $\sigma(\cdot)$ is sigmoid function normalizing both scores, then Fuse() function implements a broadcastable Harmonic Mean operation, ensuring that the link information contributes to all entity types.
Notably, spans of length 1 lack relative link score so their fused scores directly adopt the slow expert scores.

We employ Binary Cross-Entropy loss function for coherent optimization~\cite{zhang2023neural}, which independently determines whether a span belongs to a specific entity type.
Notably, Eq.~\ref{eq:norm} ensures that the final fused score $\mathbf{\tilde{h}}^{s_{ij}}$ is probabilities, we denote here as $\hat{p}_{ij}^k$. Then the loss is computed as:
\begin{gather}
\mathcal{L}_{bce} = -\sum_{i,j,k} \left[ p_{ij}^k \log \hat{p}_{ij}^k + (1 - p_{ij}^k) \log(1 - \hat{p}_{ij}^k) \right]
\label{equ_bce}
\end{gather}
where $p_{ij}^k$ is the one-hot label for span $s_{ij}$ being k-th entity type.

\subsubsection{Knowledge Retention}
We use knowledge distillation (KD) to transfer knowledge from both experts learned in previous tasks by distilling their fused scores.
Specifically, before training on a new task, we perform a one-shot prediction on the new task's data to generate pseudo labels (i.e., distilled labels) of all previously learned entity types. Than the distilled labels are used to optimize the predictions for old tasks in current step via a Bernoulli KL-divergence loss:
\begin{gather}
\mathcal{L}_{kd} = \sum_{i,j,k} \left[ \bar{p}_{ij}^k \log (\bar{p}_{ij}^k / \hat{p}_{ij}^k)
+ (1 - \bar{p}_{ij}^k) \log ((1 - \bar{p}_{ij}^k) / (1 - \hat{p}_{ij}^k)) \right]
\label{equ_kd}
\end{gather}
where $\bar{p}_{ij}^k$ is the distilled label generated by span $s_{ij}$ for the k-th entity type from old tasks. Here, $\hat{p}_{ij}^k$ is fused score similar to that in Equ. \ref{equ_bce} but specially for the old entity types.

\subsubsection{Length-Decay Negative Sampling}
In span-based models, positive spans (entities) are significantly outnumbered by negative spans. This severe imbalance leads to biased optimization during standard training. Negative sampling aims to address the imbalance by randomly removing a portion of negative spans during learning, but it may result in insufficient learning or the omission of hard negative spans. Opportunely, fast expert can provide shortcut information to reject negative spans, enabling lossless integration of negative sampling. It also mitigates the risk of overly optimizing weak links.

\begin{figure}
\centering
  \includegraphics[width=2in]{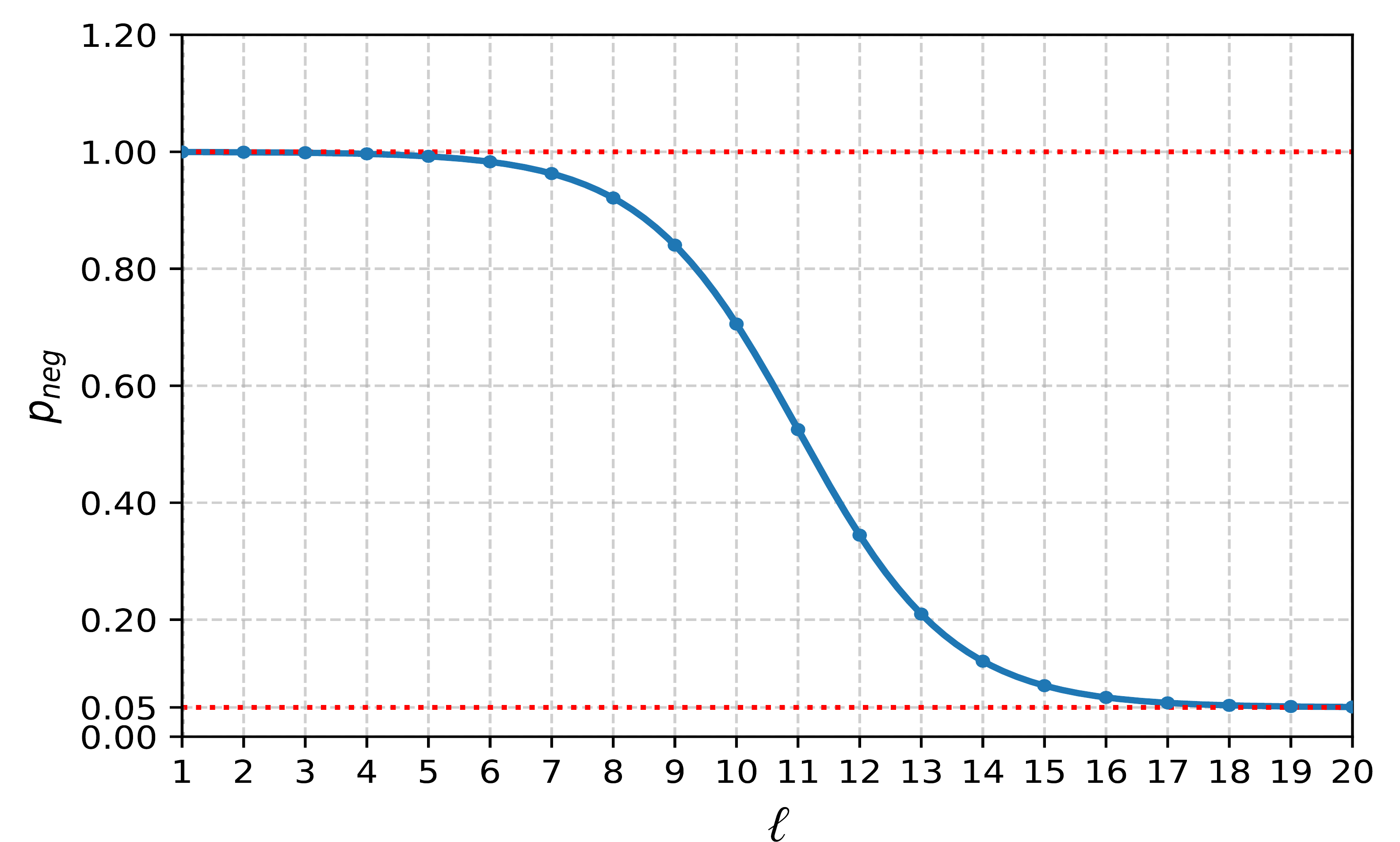}
  \vspace{-10pt}
  \caption{The Curve of designed negative sampling probability against the length of the negative span.
  }
\label{neg_sample}
\vspace{-10pt}
\end{figure}

Given the empirical observation that longer spans are less likely to be entities, we design a length-decay negative sampling strategy for our loss computation. As shown in Fig.~\ref{neg_sample}, negative sampling probabilities $P_{\text{neg}}$ corresponding to span length $\ell$ is design as:
\begin{equation}
    P_{\text{neg}}(\ell) = \eta \cdot \left( 0.95 \cdot \sigma(-0.8 \cdot \ell + 11) + 0.05 \right)
\end{equation}
where the base sampling ratio $\eta$ is set to 1 in experiments, 0.95 and 0.05 are used to scale the domain of the probability curve to the range [0.05, 1], ensuring that even overly long spans have a non-zero probability of being sampled. 0.8 controls the decay slope and 11 controls the shift of sigmoid, i.e., the length threshold at which the sampling probability begins to decline. These values and their approximate ranges were tested in preliminary experiments. The current values performed best, but other approximate settings also yielded comparable results owing to the robustness derived from the probabilistic nature. We apply negative sampling when computing both $\mathcal{L}_{bce}$ and $\mathcal{L}_{kd}$ losses, obtaining $\mathcal{L}_{bce}^{ns}$ and $\mathcal{L}_{kd}^{ns}$, respectively. Noted that sampling only apply to negative spans. The spans of entities and the spans within entities are forcibly kept. In KD, we treat span's distilled labels with values below 0.5 as negative spans.
Finally, the training loss is the weighted sum as:
\begin{equation}
    \mathcal{L} = \alpha \cdot \mathcal{L}_{bce}^{ns} + \beta \cdot \mathcal{L}_{kd}^{ns}
\end{equation}
where the weight is set to $\alpha\!=\!\beta\!=\!1$ in experiments.

\section{Experiments}\label{sec:exp}
In this section, we first introduce the datasets and metrics, followed by the training details and the selected baselines. We then present the overall performance comparisons and conduct an ablation study. Finally, we analyze the learning efficiency and provide case visualizations.

\subsection{Datasets}
In CLNER, evaluating models using datasets synthesized from traditional NER benchmarks become common practice~\cite{monaikul2021continual,zhang2023neural}, where the original dataset is partitioned into distinct segments, each serving as an individual continual learning task.
Following \cite{zhang2023neural}, we randomly split training data and retain annotations only for entity types designated for the current learning task.
The full test data is preserved for evaluation, but only entity types learned up to the current task are visible for testing.
This setup ensures the inclusion of unannotated entities from previously learned or future types, as well as purely negative sentences that contain no entities, thereby better aligning with realistic CLNER scenario.
We use two source dataset:

\indent \textbf{OntoNotes5.0-EN}~\cite{LDC2013T19},
annotated with 18 entity types, is converted by selecting 6 types to ensure adequate training samples per task and each task learn single entity type:
\textit{{\footnotesize\textsf{Organization(ORG)}, \textsf{Person(PER)}, \textsf{Geo Political Entity(GPE)}, \textsf{Date(DATE)}, \textsf{Cardinal(CARD)}, \textsf{Nationalities/Religious/Political Groups(NORP)}}}.

\indent \textbf{FewNERD}~\cite{ding2021few},
hierarchically annotated with 8 coarse-grained and 66 fine-grained entity types, is converted by assigning coarse-grained types for each task, i.e., each task learns multiple (6$\sim$12) fine-grained types. Coarse-grained types are
\textit{{\footnotesize\textsf{Location(LOC)}, \textsf{Person(PER)}, \textsf{Organization(ORG)}, \textsf{Other(OTH)}, \textsf{Product(PROD)}, \textsf{Building(BUID)}, \textsf{Art(ART)}, \textsf{Event(EVET)}}}.

\subsection{Metrics}
For each task, model is incrementally trained on its training set and evaluated on its test set using the best checkpoint based on its development set. To factor out task order impact, we follow previous works~\cite{monaikul2021continual,zhang2023neural} to use 6 permutations for OntoNotes and 4 permutations for FewNERD. Results are averaged across all permutations and the standard deviations are typically reported at the final step, when all permutations have cumulatively learned the same complete set of entities.
We report Macro-F1 across all entity types learned up to each incremental step. Macro-F1 of the final step is compared between models. Notably, FewNERD contains multiple types per task, we compute Micro-F1 for this fine-grained types within tasks due to severe imbalance, while still maintaining Macro-F1 for the final coarse-grained types. 
For all span-based models, we only keep the entity with the highest predicted score and discard the overlapping others when predicting overlapped spans.

\subsection{Implementation Details}
For contextual encoder, we use PLM \textit{bert-base-cased}~\cite{devlin2019bert} from HuggingFace
followed by a BiLSTM with an output hidden size of 400 and subsequently apply 0.2 dropout rate. (i.e., $d^{h}\!=\!400$).
We set $d^{o}\!=\!d^l\!=\!50$ on FFNs to enable simultaneous modeling of fast and slow expert as treating link computation as an additional head in span modeling.
We fine-tune all parameters using AdamW optimizer~\cite{loshchilovdecoupled} by setting learning rate $1e^{-5}$ for PLM bert and $1e^{-3}$ for the remaining, with schedule of warmup at first 200 steps followed by a cosine decay.
Sentences are tokenized by PLM and truncated to a maximum length of 512 tokens. 
We aggregate tokens within each word by using mean-pooling of BERT output to represent the final word-level output.
We maintained a consistent batch size of 32 and identical hyperparameters across all datasets, except for training 10 epochs on OntoNotes and 5 epochs on FewNERD.

\subsection{Baselines}
We compare FSE with the following representative baselines: \textbf{SeqFT} sequentially fine-tunes the model on new data without any anti-forgetting strategy based on the conventional BERT sequence tagging, providing the lower-bound.
\textbf{AddNER} and \textbf{ExtendNER} both incorporate KD under the sequence labeling framework, by using single and multiple heads for learning new tasks, respectively.
\textbf{L\&R} improves ExtendNER by reviewing and generating synthetic samples to ensure the presence of old entity types.
\textbf{ExtendNER+DLD} equips ExtendNER with decomposed logits distillation (DLD) to strengthen discriminative ability in distillation.
\textbf{SpanKL} adopts span-based paradigm with binary classification and Bernoulli KD improving the coherence of CLNER.
\textbf{SKD-NER} builds upon SpanKL with using reinforcement learning to optimize KD process.
For \textbf{LLM}, we select GPT-5 ($\mathtt{gpt}\text{-}\mathtt{5}\text{-}\mathtt{2025}\text{-}\mathtt{08}\text{-}\mathtt{05}$)~\cite{openai2025gpt5}
and Llama3.1-8B~\cite{2024llama} for comparison.
GTP-5 is directly prompted at each step to recognize the entity types intended to learn so far, while Llama is finetuned using LoRA on the same synthesized CL datasets. We also evaluate our method under a non-CL standard training regime, which uses all available training data up to the current step while retaining all annotations of entity types learned so far, as the upper-bound performance.

\begin{table}[ht]
\centering
\caption{Macro-F1 scores (\%) of different methods at each incremental step on \textbf{OntoNotes} under continual learning.}
\label{tab:ontonotes}

\renewcommand{\arraystretch}{1.2}

\resizebox{\linewidth}{!}{
\begin{tabular}{c|ccccc>{$}c<{$}}
\toprule

\multirow{2}{*}{\textbf{Method}} & \multicolumn{6}{c}{Incremental Steps} \\
\cmidrule{2-7} & Step1 & Step2 & Step3 & Step4 & Step5 & Step6 \\
\midrule
\midrule
SeqFT \cite{devlin2019bert} & 82.46 & 56.73 & 39.38 & 29.5 & 28.07 & 23.64_{\pm 4\textbf{.}2} \\
AddNER \cite{monaikul2021continual} & 82.52 & 83.90 & 84.66 & 85.02 & 85.48 & 85.03_{\pm \textbf{.}18} \\
ExtendNER \cite{monaikul2021continual} & 82.79 & 83.54 & 84.48 & 84.67 & 85.12 & 84.96_{\pm \textbf{.}15} \\
L\&R \cite{xia2022learn} & 84.02 & 82.44 & 82.02 & 81.88 & 83.22 & 84.42_{\pm \textbf{.}20} \\
ExtendNER+DLD \cite{zhang2023decomposing} & 83.97 & 85.50 & 85.94 & 86.45 & 86.58 & 86.85_{\pm \textbf{.}11} \\
SpanKL \cite{zhang2023neural} & 85.60 & 87.92 & 88.22 & 88.76 & 89.02 & 88.98_{\pm \textbf{.}10} \\
SKD-NER \cite{chen2023skd} & 85.49 & 87.98 & 88.30 & 88.67 & 89.11 & 88.92_{\pm \textbf{.}13} \\

\midrule
\textbf{LLMs} \\
GPT-5 \cite{openai2025gpt5} & 61.21 & 69.53 & 73.88 & 69.76 & 65.31 & 67.23 \\
Llama3.1 {\scriptsize (Finetuned)} & 78.30 & 65.15 & 36.62 & 25.17 & 9.10 & 14.74_{\pm 1\textbf{.}1}\\
Llama3.1 {\scriptsize (Finetuned)}$^\mathsf{m}$ & 78.30 & 76.67 & 84.93 &81.73 & 75.25 & 74.58_{\pm \textbf{.}89}\\

\midrule
\rowcolor{gray!10}
\textbf{FSE (Ours)} & 85.70 & 88.93 & 88.81 & 88.98 & 89.02 & \textbf{89.23}_{\pm \textbf{.}09}\\
\midrule
\makecell{standard non-CL\\[-1pt] (upper-bound)} & 85.70 & 89.23 & 89.27 & 89.69 & 89.75 & 90.07_{\pm \textbf{.}13} \\
\bottomrule
\end{tabular}
}
\vspace{-10pt}
\end{table}

\begin{table}[ht]
\centering
\caption{Macro-F1 scores (\%) of different methods at each incremental step on \textbf{FewNERD} under continual learning.}
\label{tab:fewnerd}
\renewcommand{\arraystretch}{1.2}
\resizebox{\linewidth}{!}{
\begin{tabular}{c|ccccccc>{$}c<{$}}

\toprule
\multirow{2}{*}{\textbf{Method}} & \multicolumn{8}{c}{Incremental Steps} \\
\cmidrule{2-9} & Step1 & Step2 & Step3 & Step4 & Step5 & Step6 & Step7 & Step8 \\
\midrule
\midrule
SeqFT \cite{devlin2019bert} & 63.87& 37.00 & 21.61& 14.69& 12.34&  11.12 & 10.79&  7.21_{\pm 2\textbf{.}8}\\
AddNER \cite{monaikul2021continual} & 64.01 & 61.32 & 60.54 & 59.43 & 58.74 & 59.32 & 60.41 & 59.32_{\pm \textbf{.}13}\\
ExtendNER \cite{monaikul2021continual} & 64.06 & 59.02 & 57.05 & 55.72 & 55.46 & 55.96 & 56.85 & 56.16_{\pm \textbf{.}08}\\
L\&R \cite{xia2022learn} & 64.06 & 59.78 & 58.07 & 55.89 & 55.45& 56.10 & 57.21 & 57.02_{\pm \textbf{.}11}\\
ExtendNER+DLD \cite{zhang2023decomposing} & 64.06 & 60.12 & 57.83 & 56.22 & 56.63 & 57.17 & 58.02 & 58.82_{\pm \textbf{.}10}\\
SpanKL \cite{zhang2023neural} & 67.81 & 64.16 & 63.62 & 62.31 & 61.67 & 62.17 & 63.24 & 62.15_{\pm \textbf{.}09} \\
SKD-NER \cite{chen2023skd} & 67.68 & 64.55 & 63.90 & 62.78 & 61.88 & 62.84 & 63.20 & 62.42_{\pm \textbf{.}11}\\

\midrule
\textbf{LLMs} \\
GPT-5 \cite{openai2025gpt5}  & 59.04 & 58.03 & 53.89 & 53.10 & 51.95 & 49.72 & 49.48 &  47.92 \\
Llama3.1 {\scriptsize (Finetuned)}  & 73.58 & 39.08 & 34.25 & 32.86 & 17.40 & 17.13 & 15.04 &  13.03_{\pm \textbf{.}84}\\
Llama3.1 {\scriptsize (Finetuned)}$^\mathsf{m}$ & 73.58 & 51.14 & 46.20 & 39.62 & 38.61 & 39.81 & 30.06&  35.57_{\pm \textbf{.}77}\\

\midrule
\rowcolor{gray!10}
\textbf{FSE (Ours)} & 66.61 & 62.43 & 62.39 & 62.64 & 62.27 & 63.78 & 64.20 & \textbf{63.69}_{\pm \textbf{.}08} \\
\midrule
\makecell{standard non-CL\\[-1pt] (upper-bound)} & 66.61 & 63.53 & 63.77 & 64.56 & 64.89 & 66.65 & 67.11& 66.83_{\pm \textbf{.}11} \\
\bottomrule
\end{tabular}
}
\vspace{-10pt}
\end{table}
\begin{table}[t]
\footnotesize
\centering
\caption{Performance of final step on two datasets evaluated by removing different components of our FSE.}
\label{tab:ablation}

\setlength{\tabcolsep}{3pt}
\renewcommand{\arraystretch}{0.9}
\resizebox{\linewidth}{!}{
\begin{tabular}{lccc}
\toprule

\multirow{2}{*}{Method} & \multicolumn{2}{c}{Dataset} &  \multirow{2}{*}{\makecell{Training Throughput\\(steps/second)}}\\
\cmidrule(lr){2-3} & OntoNotes & FewNERD \\

\midrule

\textbf{FSE (Ours)} & \textbf{89.23$_{\pm \textbf{.}09}$} & \textbf{63.69$_{\pm \textbf{.}08}$}  & $\approx11.70$ \\
$\quad$ w/o \textbf{shared} \textbf{F}ast \textbf{E}xpert & 89.02$_{\pm \textbf{.}08}$ & 63.43$_{\pm \textbf{.}06}$ & $\approx11.19$ \\
$\quad \quad$ (i.e., w/o shared information) & & &\\
$\quad$ w/o \textbf{F}ast \textbf{E}xpert & 88.77$_{\pm \textbf{.}14}$ & 62.95$_{\pm \textbf{.}11}$  & $\approx12.56$\\
$\quad$ w/o \textbf{N}egative \textbf{S}ampling & 89.00$_{\pm \textbf{.}12}$ & 63.46$_{\pm \textbf{.}09}$  & $\approx9.36$\\

\midrule
replace Fast Expert with hard pruning & & &\\

$\quad$ (pruning span with length > 10) & 86.67$_{\pm \textbf{.}14}$ & 57.46$_{\pm \textbf{.}12}$  & $\approx12.84$ \\

\bottomrule
\end{tabular}
}
\vspace{-10pt}
\end{table}

\begin{figure*}[ht]
\centering
\includegraphics[width=0.9\textwidth]{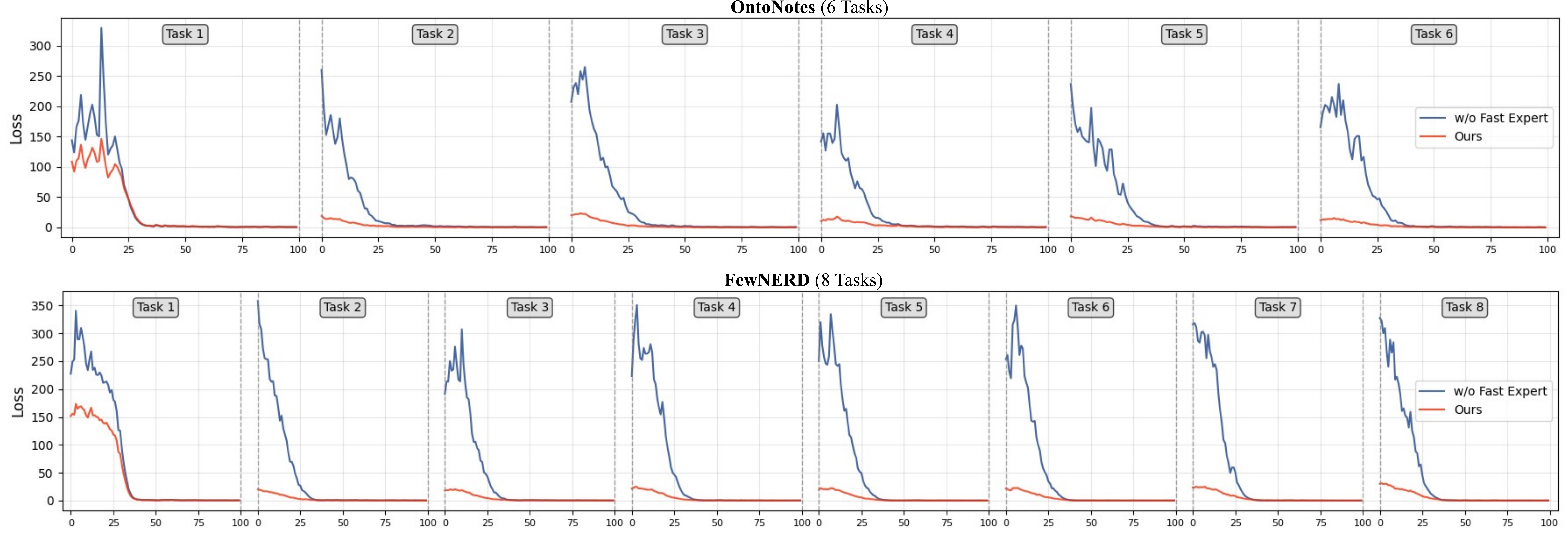}
\caption{The loss curves for the initial 100 steps of each task on both datasets show the convergence speed. The red line is FSE while blue line is FSE without fast expert. A faster decrease in loss indicates a lighter learning burden and more efficient continual learning for each task.}
\label{fig:convergence}
\vspace{-10pt}
\end{figure*}

\subsection{Overall Performance}
We report the task-wise performance after sequentially learning each tasks on two synthesized datasets of different models. The final step's metrics best represent the overall performance.

\textbf{OntoNotes}'s results are shown in Tab.~\ref{tab:ontonotes}, where FSE achieves the highest score at the final step6, demonstrating the strongest recognition capability for all accumulated entity types. \textbf{FewNERD}'s results are shown in Tab.~\ref{tab:fewnerd}, where FSE consistently maintains SoTA performance even in this challenging scenario typically learning more than 6 entity types per incremental step. The improvement of FSE is more pronounced here, indicating the potential of shared fast expert to learn sizable entity types.

\textbf{LLM} performance on both datasets reveals that neither GPT-5 nor Llama3.1 achieve optimal results, primarily due to inherent limitations of prompt-based method: As the number of predefined entity types filled in the instruction increases, it become challenging to accurately follow instruction. Also, LLMs may not fully comprehend the predefined entity type names, and the uncontrollable LLM-generated outputs occasionally lead to parsing failure, both causing incorrect extractions.
This is evidenced by the poorer results on FewNERD which contains total 66 entity types.
Notably, original result of finetuned Llama encounters train-test inconsistency since the instruction are filled with single task entity types in training but involve multiple tasks in prediction. We thereby perform multiple predictions (results with $^{\boldsymbol{\mathsf{m}}}$), each focusing on a single task entities, and then aggregate them. This finally improve the performance but still remain suboptimal.

\begin{figure*}[h]
\centering
\includegraphics[width=0.9\textwidth]{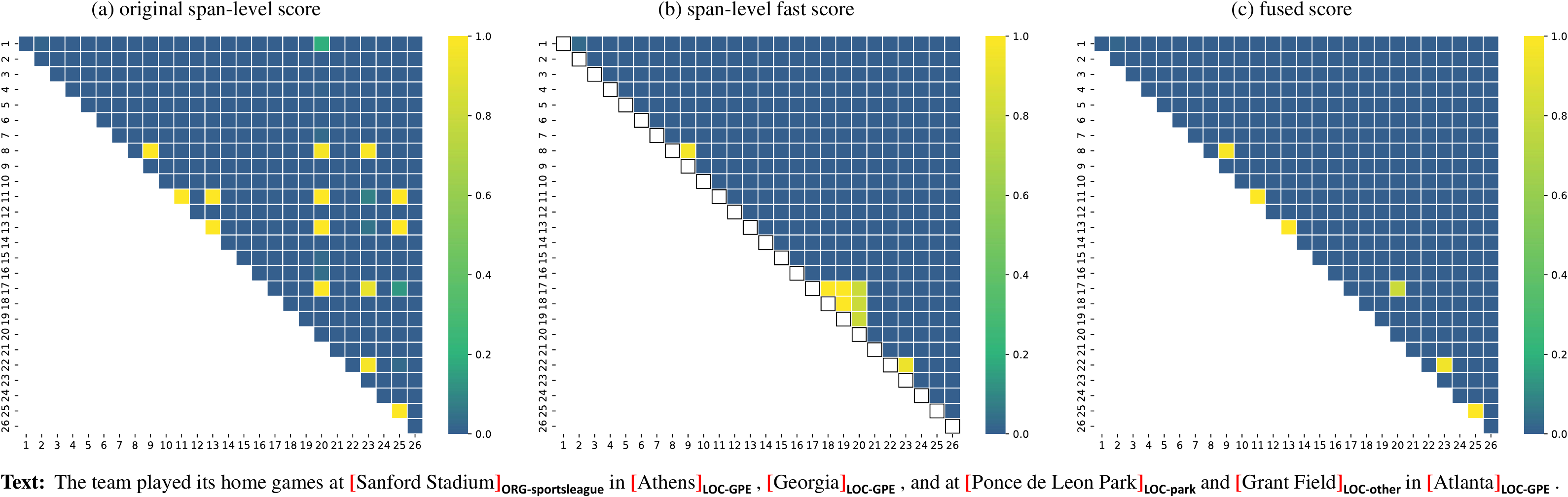}
\caption{Values probed from the span matrix view of a sampled test sentence on FewNERD, with coarse-grained types (\texttt{ORG}, \texttt{PER}), using FSE final step model.
(a) Loose span scores from span modeling by slow expert, showing reduced learning burden. (b) Span-level fast scores derived from link scores by fast expert, are shared up to the last step. (Hollow boxes denote spans of length 1.) (c) Fused scores between both experts, where fast expert can refine the 'arbitrary' results from the slow expert to accurately recognize entities.
}
\label{tab:probe}
\vspace{-10pt}
\end{figure*}

\subsection{Ablation Study}
We conduct ablation study to verify the effectiveness of each design in FSE. The ablated variants includes: \textbf{w/o shared FE} removes the sharing mechanism of the fast expert across tasks, meaning each task independently initializes a new fast expert, which can exclusively evaluate the effect of using shared information. \textbf{w/o FE} removes the entire fast expert module, retaining only the slow expert for span modeling. \textbf{w/o NS} removes the negative sampling strategy.
We report the final step result as the overall performance.

As shown in Tab.~\ref{tab:ablation}, removing different components of FSE results in a decline in overall performance. Non-shared fast expert leads to a performance drop, which indicates that the sharing mechanism of the fast expert is useful as it allows each task to leverage the previously learned fast expert, utilizing the shared information across tasks. Removing the entire fast expert module results in a more significant performance degradation, demonstrating the advantage of fast expert in reducing the learning burden to comprehensively enhance model performance. Results without negative sampling strategy also confirm its contribution to FSE. The training throughput on H800 when using different components are also reported, showing that fast expert does not significantly slow down the hardware training speed due to its lightweight architecture and the parallel implementation of fast score propagation. We also evaluate a hard pruning variant that replaces the neural-learning based link and fast score of the fast expert with a heuristically set length threshold to prune overly long spans. The threshold is set to 10 to consider avoiding unnecessary loss of long entities appearing in both corpora. The results show that this rigid strategy reduces performance, as it lacks the flexibility in handling long entities and the plasticity in continual learning.

\subsection{In-Depth Analysis}
\textbf{Learning Efficiency.}
FSE is expected to reduce the learning burden for each entity type by rapidly soft filtering out the majority of candidate spans using neural-based fast scores. Here, we plot the loss curves during the initial 100 training steps of each task to verify whether FSE improves the convergence rate, where a faster convergence rate implies easier learning.
Fig.~\ref{fig:convergence} shows the loss curves for each training tasks on OntoNotes and FewNERD, respectively. Compared to removing the fast expert, FSE accelerates learning per task, as reflected in faster loss convergence. It becomes more pronounced after the first task, indicating that the learned fast expert is successfully shared to promotes the newcome tasks. Notably, the counterpart (i.e., w/o fast expert, having non-weak performance in the ablation study) essentially represents existing competitive span-based method (e.g., SpanKL, with the same span modeling architecture). The comparison results thereby reveal a defect in current strong span-based approaches. Reducing the learning difficulty of each task enable models to better handle the incoming tasks, which explains the improvement during continual learning. Moreover, the learning efficiency can help save training time by allowing fewer training epochs in practice.

\textbf{Visualization.}
To intuitively verify the functionality of both experts, we probe the values in the span matrix for visualization.
The span matrix is defined as the scores of each span arranged in a matrix perspective. 
Here, we consider three matrices underlying FSE: (a) the original span score matrix derived from slow experts, (b) the fast score matrix derived from the fast expert, and (c) the resulting fused score between them.
Noted that we aggregate the span scores of all entity types by max-pooling for simultaneous display.
Fig.~\ref{tab:probe} visualizes a sample randomly selected from test set of FewNERD.

Evidently, the presence of considerable number of high scores in (a) indicate that the slow expert can afford some mistake in span modeling, as these raw scores will ultimately be correctly deactivated by the fast score of fast expert.
This robust span modeling allows the slow expert to focus only on the spans remained in link-enabled area (the high scores area in (b)), thereby enhancing its learning capacity.

\section{Conclusion}\label{sec:con}
\vspace{-2pt}
This paper propose a fast-slow experts enhanced span-based model FSE for CLNER.
The fast expert mitigates task interference by modeling shared link information between adjacent tokens to filter improbable spans across tasks, improving stability.
The slow expert efficiently focuses on the remaining spans to reduce learning burden, improving plasticity.
The tailored length-decay negative sampling strategy applied to both experts handles the span imbalance issue.
Comprehensive experiments on synthetic datasets derived from real-world corpora OntoNotes and FewNERD confirm that FSE achieves superior performance compared to baseline including LLM-based competitor. The ablation studies validating the contribution of each component. The loss curve verifies the accelerated convergence per task, and the visualized results validate the expected functionality of both experts. 
Future work will explore more sophisticated link modeling mechanisms and extend this framework to broader information extraction tasks under continual learning.

{\small

\bibliographystyle{elsarticle-num-names}
\bibliography{main}

}
\end{document}